\documentclass[10pt]{article}
\usepackage[accepted]{rlj} 

\usepackage{amssymb}
\usepackage{mathtools}
\usepackage{graphicx}
\usepackage{subcaption}
\usepackage[space]{grffile}
\usepackage{url}
\usepackage{booktabs}
\usepackage{tabularx}
\usepackage{placeins}
\usepackage{float}
\usepackage{listings} 

\title{Hierarchical Control in Multi-Agent Games: LLM-based Planning and RL Execution}

\setrunningtitle{Hierarchical Control in Multi-Agent Games: LLM-based Planning and RL Execution}

\author{Jannik Hösch\textsuperscript{1,2,*}, Alessandro Sestini\textsuperscript{1}, Florian Fuchs\textsuperscript{1}, Amir Baghi\textsuperscript{1}, Joakim Bergdahl\textsuperscript{1}, Ioland Leite\textsuperscript{2}, Konrad Tollmar\textsuperscript{1}, Jean-Philippe Barrette-LaPierre\textsuperscript{1}, Linus Gisslén\textsuperscript{1,*}}

\emails{\{jhosch, asestini, ffuchs, abaghi, jbergdahl, ktollmar, jbarrettelapierre, lgisslen\}@ea.com}

\affiliations{$^{1}$\textbf{Electronic Arts (EA)}\\
$^{2}$\textbf{KTH Royal Institute of Technology, Stockholm, Sweden}\\
$^*$Corresponding authors}

\begin{document}

\maketitle

\begin{abstract}
Reinforcement learning (RL) has achieved strong performance in sequential decision-making, yet scaling to complex multi-agent environments remains challenging due to sparse rewards, large state-action spaces, and the difficulty of learning coordinated strategies. We propose a hierarchical architecture where a pretrained large language model (LLM) acts as a centralized strategic controller that selects among specialized RL skill policies for a team of agents, while RL policies handle reactive low-level execution. We evaluate this hybrid system in a competitive 2v2 King of the Hill environment against behavior tree (BT) and \emph{``Flat''} RL (end-to-end training without skill decomposition) baselines. The LLM+RL system achieves task performance statistically equivalent to hand-crafted BT (46.4\% vs 51.5\% win rate, $p=0.103$) while both significantly outperform Flat RL trained without skill decomposition. A user study ($n=15$) reveals that 60\% of participants perceive LLM+RL agents as the most human-like ($p=0.027$), citing behavioral adaptability and tactical variability. These results demonstrate that pretrained LLM reasoning can effectively orchestrate pretrained RL skills, achieving competitive multi-agent coordination and superior perceived believability without manual rule engineering.
\end{abstract}

\section{Introduction}
\label{sec:introduction}
Reinforcement learning (RL) has achieved remarkable results in sequential decision-making tasks, from board games \citep{silver2017} to robotics \citep{haarnoja2019}. However, applying RL in complex multi-agent environments remains fundamentally challenging \citep{zhang2021}. Tasks often involve partial observability, large state-action spaces, sparse rewards, and long temporal dependencies. In multi-agent scenarios, agents must additionally coordinate and adapt to non-stationary dynamics induced by other learning agents.

Hierarchical reinforcement learning (HRL) addresses some of these challenges by decomposing tasks into reusable skills (options) that operate over multiple timesteps \citep{sutton1999between}. This temporal abstraction reduces the effective decision horizon and improves credit assignment. However, HRL introduces its own challenges: discovering useful skills automatically and learning effective high-level policies that select among skills remain difficult \citep{bacon2016,klissarov2025}. Moreover, training hierarchical policies is often unstable and sample-inefficient compared to end-to-end approaches \citep{nachum2018}.

Large language models (LLMs) have demonstrated strong capabilities in reasoning, planning, and knowledge representation \citep{openai2024, wei2022}. This motivates hybrid architectures that leverage LLMs for high-level reasoning while RL agents handle low-level execution. In such architectures, LLMs can guide strategy, decompose tasks, and coordinate multiple agents without requiring manually engineered symbolic logic. By serving as high-level controllers, LLMs have the potential to improve both decision quality and the perceived believability of autonomous agents.

We propose a two-layer hierarchical architecture where a pretrained LLM acts as a centralized meta-controller, selecting among specialized RL skill policies for a team of agents based on the global game state. The LLM operates at a slower timescale, providing strategic coordination, while RL policies execute at high frequency ensuring reactive control. To evaluate this method, we created a bespoke competitive 2v2 King of the Hill environment to compare against a behavior tree and a \emph{``Flat''} RL baseline. The Flat RL baseline uses a single PPO policy that learns all tactical behaviors end-to-end without skill decomposition, with each agent acting on local observations.

Our contributions are: (1) a hierarchical LLM+RL architecture for competitive multi-agent coordination that combines pretrained LLM reasoning with pretrained RL skill policies, (2) empirical evidence that this architecture achieves task performance statistically equivalent to hand-crafted behavior trees while significantly outperforming Flat RL, and (3) a user study demonstrating that LLM-guided agents are perceived as significantly more human-like by players, with 60\% selecting them as most resembling competent human opponents.

\section{Related Work}
\label{sec:related}

\paragraph{Traditional Game AI.}
Classical game AI approaches range from reactive systems like behavior trees (BTs), which offer modular control but require substantial manual authoring \citep{colledanchise2018,mcclarron2016}, to symbolic planners \citep{orkin2006,ghallab2004}, which enable proactive reasoning but demand extensive domain modeling and struggle with uncertainty.

\paragraph{Multi-Agent RL.}
However, multi-agent settings introduce challenges such as non-stationarity and non-unique learning
goals \citep{zhang2021}, complicating credit assignment and making traditional end-to-end RL approaches struggle when training for complex coordination \citep{lowe2020}.

\paragraph{Hierarchical RL.}
HRL addresses these challenges through temporal abstraction, decomposing tasks into reusable skills (options) \citep{sutton1999between} that shorten decision horizons and improve credit assignment. RL has successfully been used in previous studies as a high-level strategic component to control sub-policies in games \citep{bergdahl2024reinforcement}. However, automatic skill discovery and hierarchical policy learning remain challenging and often sample-inefficient compared to end-to-end approaches \citep{bacon2016,klissarov2025,nachum2018}. When domain structure is well understood, manually specified skills offer a pragmatic alternative.

\paragraph{LLMs for Planning.}
LLMs exhibit emergent capabilities in zero-shot reasoning \citep{kojima2023,naveed2024} and task planning \citep{huang2024,yao2023}. However, their inference latency and lack of grounded physical representations \citep{ji2023} make them unsuitable for low-level reactive control, motivating their use for high-level strategic reasoning at slower timescales. Recent work explores LLMs as components in RL systems: as reward designers \citep{ma2024,afonso2025self} and task decomposers \citep{yang2025,cao2025}.

\paragraph{Hierarchical LLM-RL Systems.}
Recent work combines these directions by using LLMs as high-level decision-makers in hierarchical RL architectures, where the LLM performs strategic reasoning while low-level policies execute concrete actions. For single-agent embodied tasks, SayCan \citep{ahn2022} combines LLM planning with learned affordance models for robotic manipulation, while Voyager \citep{wang2023} demonstrates autonomous skill library construction in Minecraft. In multi-agent settings, L2M2 \citep{geng2025} uses LLMs to generate navigation subgoals for cooperative MARL, reducing training samples by 80\% compared to end-to-end approaches. YOLO-MARL \citep{zhuang2025} generates episode-level strategies for cooperative tasks, while recent frameworks apply LLMs to multi-robot task allocation \citep{kawabe2026hierarchical} and graph-based coordination \citep{jia2025enhancing}.

These works demonstrate effective hierarchical separation between strategic reasoning and reactive control, but primarily focus on cooperative scenarios with evaluation limited to task performance. Our work differs in three key ways. First, we address competitive multi-agent games where agents must balance team coordination, individual survival, and combat to outperform adversarial opponents. Second, we leverage pretrained components at both levels (RL skills and LLM), avoiding the sample complexity of learning hierarchies. Third, we evaluate both quantitative performance and human perception via a user study, addressing a critical dimension for game AI where player experience matters alongside optimization metrics.

\section{Method}
\label{sec:method}
This section presents the proposed hierarchical architecture, the evaluation environment, and baseline methods.

\subsection{Architecture}
\begin{figure}[ht]
    \centering
    \includegraphics[width=0.6\textwidth]{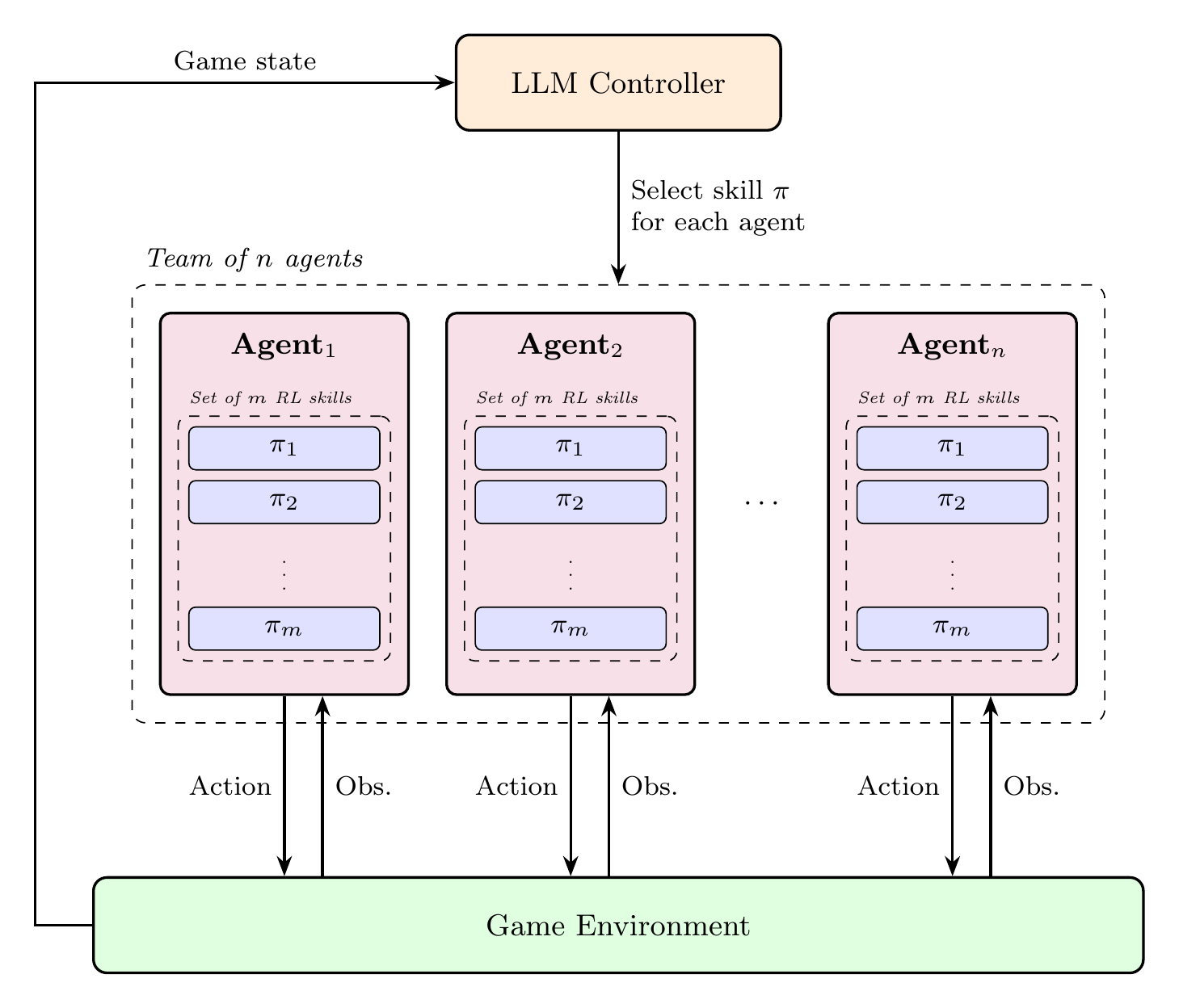}
    \caption{Architecture of the hierarchical LLM-RL system. A centralized LLM selects one of $m$ skill policies for each of $n$ agents based on global game state. Each agent executes its assigned skill, mapping local observations to actions.}
    \label{fig:architecture}
\end{figure}

The proposed architecture follows a hierarchical RL paradigm, decoupling strategic reasoning from skill execution through a two-layer hierarchy (Figure~\ref{fig:architecture}):

\paragraph{Strategic layer (LLM):} A pretrained LLM acts as a centralized meta-controller for a team of $n$ agents. At each decision step, it observes the global game state and selects for each agent one of $m$ available skill policies. The LLM operates at a slower timescale, issuing decisions based on semantic reasoning over natural-language state representations.
\paragraph{Execution layer (RL):} Each of the $n$ agents executes a skill policy selected by the LLM, mapping local observations to low-level actions at high frequency. All agents share access to the same set of $m$ pretrained RL skill policies, each specialized for a specific tactical objective.

The architecture combines state representation asymmetry (global vs. local observations) and temporal abstraction (different decision frequencies) to leverage complementary strengths: the LLM provides semantic reasoning over global state at lower frequencies for team coordination, while RL policies enable reactive control over local observations at high frequencies.

\subsection{Environment}
\begin{figure}[ht]
    \centering
    \includegraphics[width=0.6\textwidth]{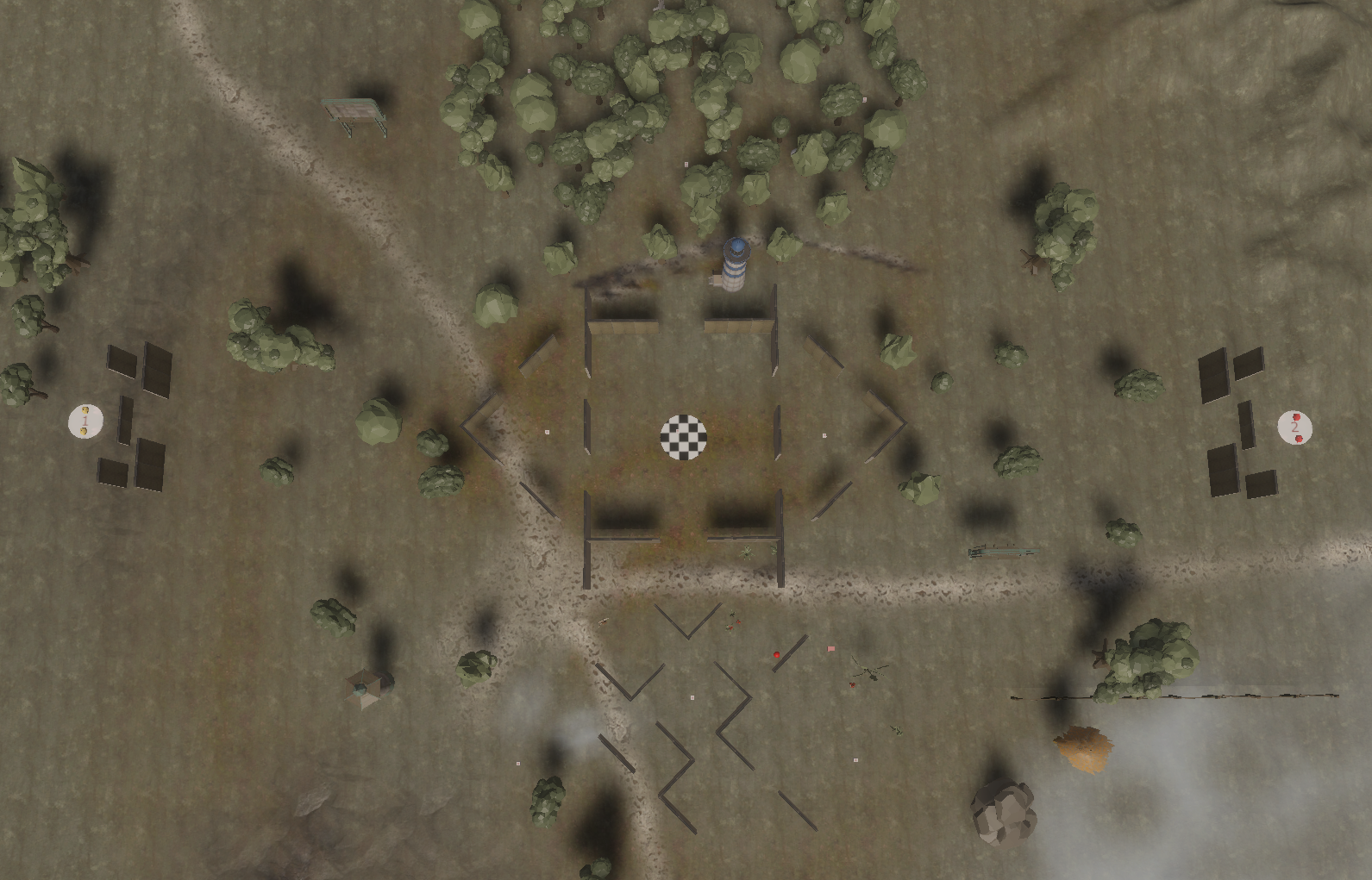}
    \caption{Top-down view of the 2v2 King of the Hill game arena in Unity. The capture zone is represented by the checkerboard circle; home bases are situated on the left and right sides of the image, respectively.}
    \label{fig:environment}
\end{figure}

To evaluate the method, we created an environment that is playable by both humans and agents, features strategic elements, and supports collaboration. We created a bespoke 2v2 King of the Hill arena built in Unity~6 (Figure~\ref{fig:environment}). Two teams of two agents compete to capture and hold a randomly placed goal zone. The team that accumulates 8 seconds of zone occupation wins. When an agent is eliminated, its team loses 50\% of goal progress, and the agent respawns. Health pickups restore full health on contact. Episodes end in a draw after 60 seconds if no team wins before that. The simulation runs at 50 Hz (0.02s timestep).

\paragraph{State Space.} Each agent observes a 20-dimensional normalized vector relative to its own position and orientation (Table~\ref{tab:obsSpace}). The observation comprises distances and angles to game elements (goal, enemies, ally, health pickups), agent state (health, time remaining), and 7 forward ray-cast distance measurements. All spatial features are normalized to [0,1] (-1 if not visible) for distances and [-1,1] for angles.

\paragraph{Action Space.} The action space is hybrid, combining three continuous actions (forward/strafe movement, rotation) in $[-1,1]^3$ and two discrete binary actions (jump, fire) in $\{0,1\}^2$ (Table~\ref{tab:actionSpace}).

\subsection{Implementation}
The concrete system comprises two agents ($n=2$) controlled by a single centralized LLM (Gemma~3 27B \citep{gemma3} via Ollama), which assigns each agent to one of four pretrained RL skill policies ($m=4$) with distinct tactical objectives::

\begin{itemize}
    \item \textbf{Navigate:} Move toward the goal zone.
    \item \textbf{Combat:} Engage and eliminate visible enemies.
    \item \textbf{Secure:} Defend the goal zone from inside.
    \item \textbf{Retreat:} Escape to health pickups when critically damaged.
\end{itemize}

\paragraph{Strategic layer (LLM). }
The LLM operates at approximately 0.5-second intervals (2 Hz), processing a two-part prompt at each decision step (Appendix~\ref{app:llmPrompt}). The static system prompt (Listing~\ref{lst:systemPrompt}) encodes task-invariant information: game rules, skill usage guidelines, team coordination principles, and output format requirements. The dynamic user prompt provides the current game state (Table~\ref{tab:llmObs}), including per-agent observations (enemy visibility and distance, health level, distances to goal zone and health pickups, current skill assignment) and team-level state (goal progress, time remaining, inter-agent distance). State variables are discretized into categorical labels (e.g., health: HIGH/MEDIUM/LOW) rather than continuous values to enable semantic reasoning over tactical situations. The generation temperature is set to 0.1 for near-deterministic outputs.

\paragraph{Execution layer (RL). } Each RL skill policy operates at 12.5 Hz (every 4 physics frames), receiving the observation vector and producing actions. All skills share the same observation and action spaces but are trained with different reward functions (Appendix~\ref{app:rewardValues}). Training is conducted independently for each skill via PPO \citep{schulman2017} for 10 million steps using Unity ML-Agents \citep{juliani2020} in dedicated training scenes designed to isolate specific tactical objectives with simplified heuristic opponent behaviors. Reward functions are shaped to guide each skill toward its strategic purpose: Navigate rewards distance-to-goal reduction with no combat incentive; Combat rewards damage dealt and angle-to-enemy alignment to maintain engagement; Secure combines zone occupation and combat rewards for capturing and defending; Retreat prioritizes survival through damage penalties and strong health-pickup collection rewards. Across all skills, weapon-fire and jump penalties suppress action spam.

\subsection{Baselines}
\paragraph{Behavior Tree (BT).} 
A hand-crafted priority tree with rule-based controls implementing a fixed tactical hierarchy. The BT has access to the same global information as the LLM but operates on a subset determined by its hard-coded rules: health thresholds, goal zone proximity, enemy visibility, and shared blackboard state. Each agent evaluates conditions in priority order: retreat (low health), secure/defend (in/near goal zone), engage enemies (when visible), then navigate. Team coordination uses a shared blackboard for role exclusivity: when one agent activates Secure, the other provides perimeter defense via Combat or Navigate.

\paragraph{Flat RL.} A single PPO policy trained per-agent end-to-end on the full task without skill decomposition. Each agent receives local observations and executes the same policy, learning all tactical behaviors within one unified network. Training used a two-stage curriculum to avoid generalization failures of pure self-play, where agents can converge on policies that exploit each other but perform poorly against diverse opponents: 15M steps of self-play established basic behaviors, followed by 15M steps against a simple heuristic team that navigates to the goal and shoots visible enemies but lacks coordination and retreat behaviors. The reward function balances combat effectiveness, navigation, zone capture, and survival, including weapon-fire and jump penalties to discourage action spam (Appendix~\ref{app:rewardValues}).

\subsection{Evaluation}
We employ two complementary evaluations: (1) a performance evaluation where each agent type is matched against every other over 1,000 episodes per matchup (3,000 total), measuring win rates, agent-level metrics (kills, deaths, damage dealt/taken, shots fired/hit, health pickups collected), and skill transition patterns (which skills are when selected, and under what game state conditions); and (2) a user study ($n=15$) where participants play against each agent type in a within-subjects design, rating opponents on challenge, skill, coordination, enjoyment, and human-likeness on 5-point Likert scales.

\section{Results}
\label{sec:results}

\subsection{Performance Evaluation}
\begin{figure}[ht]
    \centering
    \includegraphics[width=0.85\textwidth]{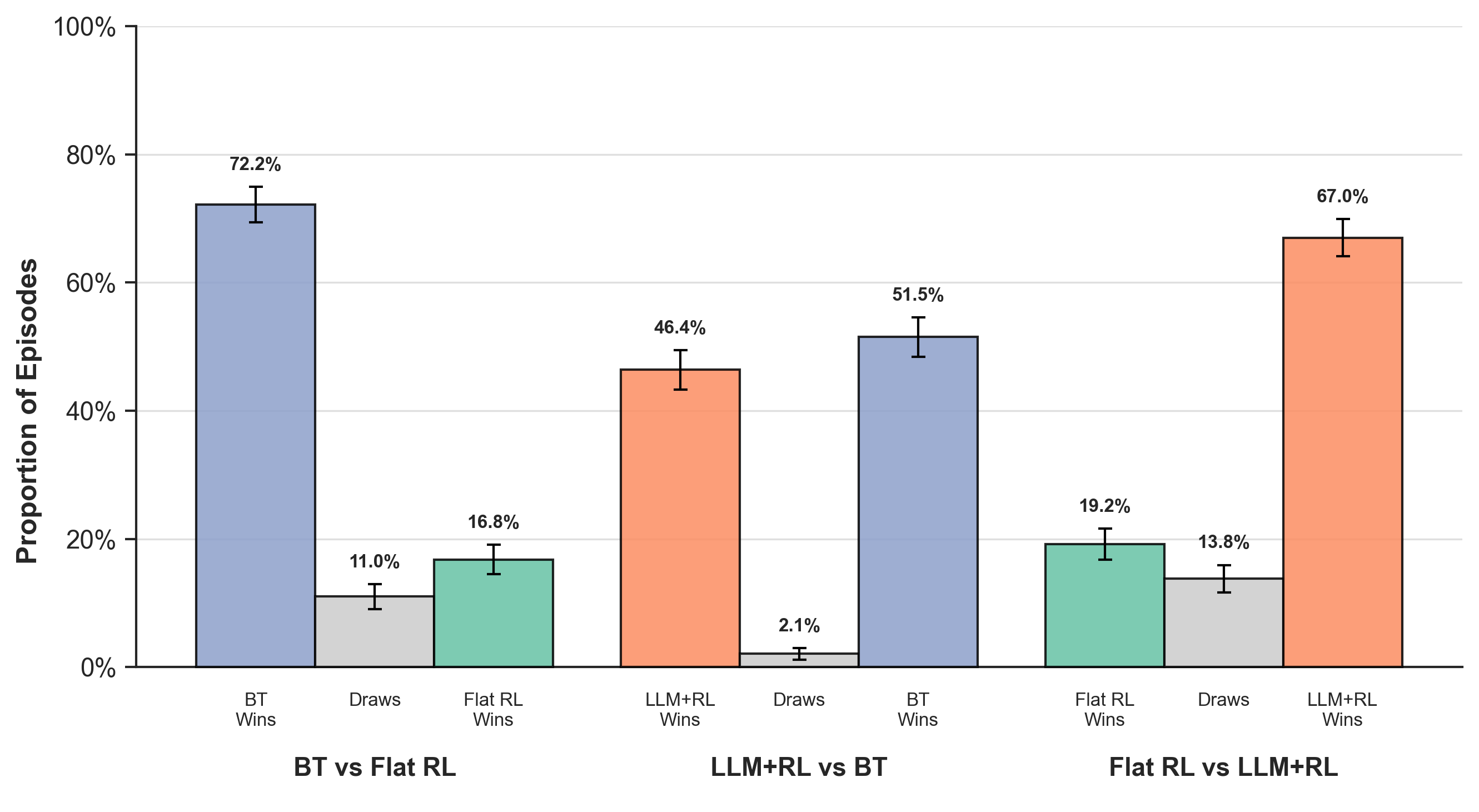}
    \caption{Episode outcome distribution for three pairwise matchups ($n=1{,}000$ each). Error bars indicate 95\% Wilson score confidence intervals. Both BT and LLM+RL achieved dominant win rates against Flat RL: BT won 72.2\% (Flat RL: 16.8\%, draws: 11.0\%), while LLM+RL won 67.0\% (Flat RL: 19.2\%, draws: 13.8\%). The BT vs LLM+RL matchup was considerably closer, with BT winning 51.5\% and LLM+RL 46.4\% (draws: 2.1\%). }
    \label{fig:winRates}
\end{figure}

\paragraph{Win Rates.}
LLM+RL achieves performance statistically equivalent to the BT while having a significantly higher win rate than Flat RL (Figure~\ref{fig:winRates}). Statistical testing (z-tests for proportions) confirmed both BT and LLM+RL significantly outperformed Flat RL ($p<0.001$), while BT's advantage over LLM+RL was not significant ($p=0.103$).

\paragraph{Agent-Level Metrics.}
While win rates show which teams won, agent-level metrics reveal the individual behaviors underlying those outcomes. Figure~\ref{fig:agentMetricsAggregate} (Appendix~\ref{app:agentLevelMetrics}) presents the aggregate distributions across all episodes, showing overall agent characteristics (K/D ratio, health pickups) and combat behavior patterns.

Flat RL demonstrated the highest Kill/Death (K/D) ratio of 2.074, achieving approximately two eliminations for every death (Figure~\ref{fig:agentMetricsAggregate}a). LLM+RL maintained an almost balanced K/D ratio of 0.901, achieving roughly equal kills and deaths, while BT had the lowest K/D ratio of 0.406, dying approximately 2.5 times per kill. While Flat RL achieved the highest K/D ratio, it recorded the lowest team-level win rates against both BT and LLM+RL, emphasizing the importance of coordination in our scenario.

Combat patterns revealed distinct strategic trade-offs (Figure~\ref{fig:agentMetricsAggregate}c--f). Flat RL adopted a combat-focused strategy: firing the most shots (median $= 87$) with moderate accuracy (56\%), dealing high damage (240) while sustaining low damage (130). BT exhibited the opposite pattern: firing few shots (26) with very high accuracy (96\%), yet dealt the least damage (120) and sustained the most damage (195). LLM+RL fell between these extremes, firing many shots (81) with low accuracy (32\%), resulting in moderate damage output (135) and moderate damage sustained (175).

LLM+RL collected the most health pickups per episode (mean $= 0.57 \pm 0.02$), followed by BT ($0.35 \pm 0.02$), while Flat RL rarely collected health ($0.08 \pm 0.01$) (Figure~\ref{fig:agentMetricsAggregate}b). This pattern reflects different survival strategies: LLM+RL and BT relied on health recovery after taking damage, while Flat RL focused on damage avoidance.

\FloatBarrier
\subsection{User Study}
To evaluate perceived agent behavior, 15 participants played against each agent type in a within-subjects design. 

\paragraph{Human-Likeness.}
LLM+RL was selected as most human-like by 60\% of participants ($n=9$), Flat RL by 33.3\% ($n=5$), and BT by 6.7\% ($n=1$). A chi-square test confirmed the distribution differed significantly from uniform ($p=0.027$, Cram\'er's $V=0.69$). Participants emphasized behavioral adaptability (\textit{"any time I got the feeling that they adapted their behavior to mine it felt more human-like"}), tactical diversity (\textit{"Multiple actions at the same time"}), and strategic reasoning (\textit{"using the health pick ups when low on health"}), noting the agents \textit{"seemed to match how I played the game"}. Complete participant responses are provided in Appendix~\ref{app:userStudyResponses}.

\begin{figure}[ht]
    \centering
    \includegraphics[width=\textwidth]{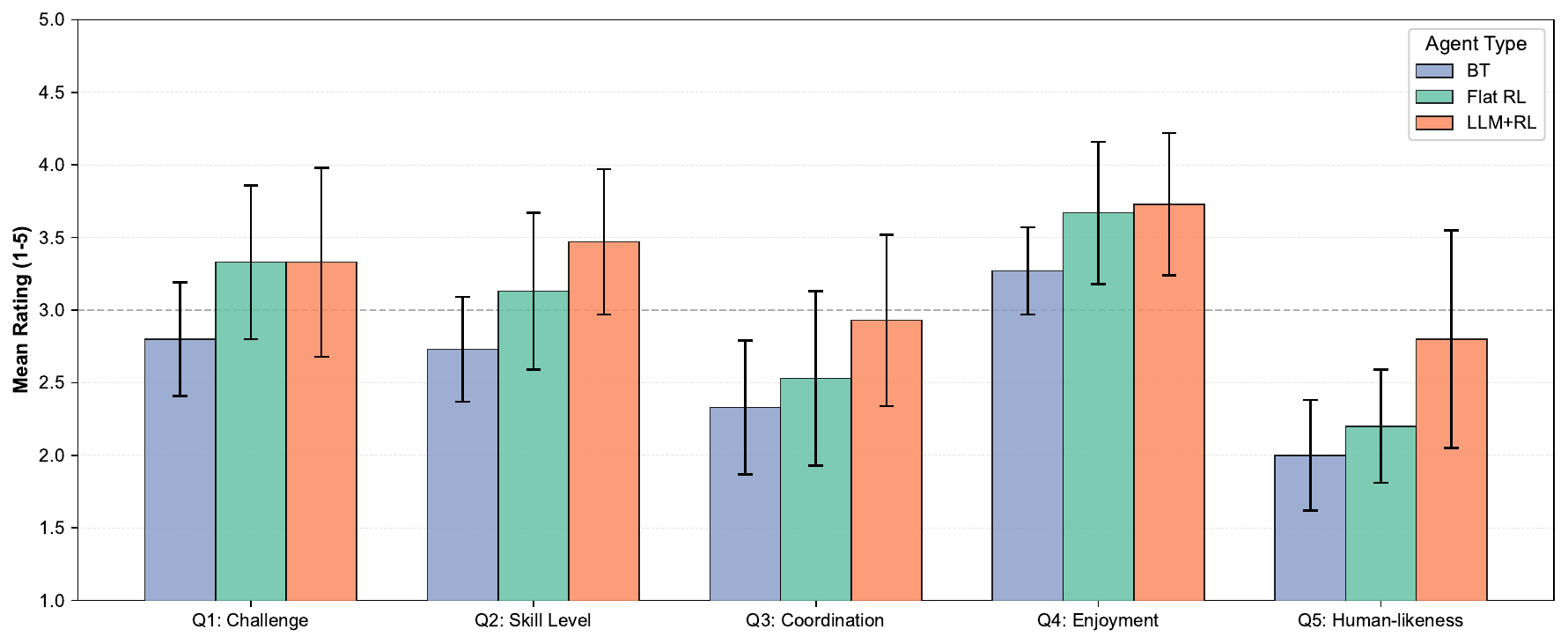}
    \caption{Post-block subjective ratings ($n=15$). Bars show means on 5-point Likert scales. Error bars: 95\% confidence intervals. Dashed line marks neutral (3.0).}
    \label{fig:likert}
\end{figure}

\paragraph{Subjective Ratings.}
LLM+RL received the highest mean ratings on four of five dimensions (Skill, Coordination, Enjoyment, Human-likeness), tying with Flat RL on Challenge (Figure~\ref{fig:likert}). BT received the lowest ratings across all dimensions. The largest difference appeared on Human-likeness, where LLM+RL exceeded BT by 0.80 points, though no agent reached a more than neutral rating (3.0). 

\begin{figure}[ht]
    \centering
    \includegraphics[width=0.8\textwidth]{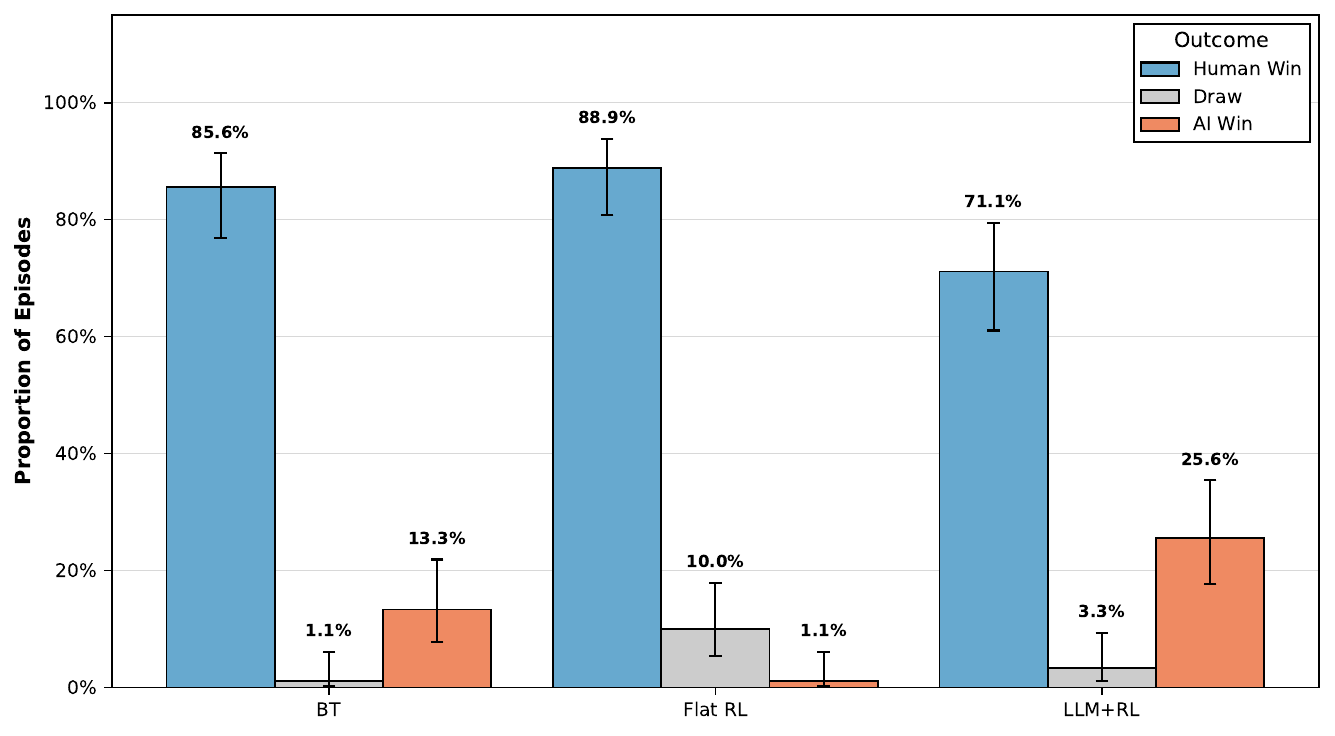}
    \caption{Episode outcomes for human teams vs each AI type ($n=90$ per opponent). LLM+RL achieved 25.6\% AI wins compared to BT (13.3\%) and Flat RL (1.1\%).}    
    \label{fig:userstudywin}
\end{figure}

\paragraph{Objective Performance Against Humans.}
To complement subjective ratings, we analyzed objective game outcomes and combat metrics from the episodes of the user study. Figure~\ref{fig:userstudywin} shows the distribution of episode outcomes for each opponent type. LLM+RL achieved the highest AI win rate against human teams (25.6\%), compared to BT (13.3\%) and Flat RL (1.1\%). Human teams won the majority of episodes against all agent types, but the margins differed significantly ($p<0.001$, Cram\'er's $V=0.294$).

\FloatBarrier
\subsection{Skill Analysis}
To understand how the LLM's decision making compares to rule-based control by the BT, we analyze its skill selection patterns. While both architectures use the same four skills, LLM+RL allocates more time to Combat (36.2\% vs 24.0\%) and less to Navigate (40.8\% vs 49.9\%). 

Temporal skill analysis (Appendix~\ref{app:llmAnalysis}) reveals strategic adaptation: both agents start with Navigate, transition to Combat/Secure mid-episode, and increase Retreat as damage accumulates (Figure~\ref{fig:skillOverTime}). However, LLM+RL shifts more decisively toward sustained zone control after initial positioning. This context-dependency is statistically confirmed: Retreat dominates at low health, Secure when in goal zone (Figure~\ref{fig:decisionContext}). LLM+RL's higher transition entropy ($H=3.19$ vs $H=2.47$ bits, Figure~\ref{fig:skillTransitions}) indicates varied transitions rather than stereotyped loops. These patterns demonstrate adaptive decision-making that explains both performance and perceived human-likeness.

\section{Conclusion, Limitations, \& Future Work}
\label{sec:conclusion}

We investigated whether pretrained LLMs can serve as effective high-level controllers in hierarchical RL systems for multi-agent coordination. Our two-layer architecture, where a Gemma~3 27B LLM selects among four pretrained RL skill policies in a competitive 2v2 environment, provides evidence addressing two questions. First, LLM-guided skill selection achieves competitive task performance, matching hand-crafted BT while significantly outperforming Flat RL. The hierarchical decomposition addresses the credit assignment challenges through temporal abstraction, enabling the LLM to balance combat, zone capture, and survival through context-dependent skill selection. Second, LLM-guided agents are perceived as significantly more human-like: 60\% of participants selected LLM+RL as the most human-like opponent, valuing behavioral adaptability and unpredictability over raw combat metrics. None of the methods reached neutral (3.0) on the Human-likeness scale, with LLM+RL achieving the highest rating of 2.8, potentially reflecting an inherent bias where participants anchor expectations below human-level when aware they are playing against AI.

The central finding is that pretrained LLM reasoning can orchestrate pretrained RL skills to achieve both competitive multi-agent coordination and superior player experience without manual rule engineering. The LLM addresses the option discovery problem by leveraging pretrained knowledge for semantic task decomposition, inferring strategic phase progression, and context-dependent skill selection without hard-coded logic. Practically, the natural-language interface reduces development complexity: designers specify skills while delegating strategic reasoning to the LLM.

Limitations include evaluation in a single environment with one LLM model (Gemma~3 27B), a modest user study sample ($n=15$), and a manually designed prompt without systematic ablation. Reasons why the architecture did not outperform the BT could be due to the LLM's lack of skill execution awareness, occasional premature switching, and RL policies encountering out-of-distribution states after transitions.

Future work spans three directions: domain generalization, technical refinement, and architectural extension. First, evaluating across diverse scenarios would test generalizability, including cooperative tasks, larger teams, and different game modes. Beyond games, multi-robot systems present a natural application domain where the natural-language interface could enable non-expert operators to issue high-level directives. Second, improving the LLM component through systematic prompt ablation, evaluating different model sizes for reasoning-latency trade-offs, and exploring event-based architectures that respond to critical state changes rather than fixed intervals. Third, extending skill representation from parameterized skills (specifying which enemy to engage, where to navigate) to fully automated skill discovery. One direction could integrate multiple LLM roles: the LLM decomposes objectives into subgoals, generates reward functions following the reward designer paradigm \citep{ma2024, afonso2025self}, trains skill policies, and orchestrates their execution, closing the loop from task decomposition to hierarchical coordination in a fully LLM-driven pipeline requiring only the environment and objective as input.

\bibliography{main}
\bibliographystyle{rlj}

\cleardoublepage

\beginSupplementaryMaterials

\appendix

\section{Observation and Action Spaces}
\label{app:obsActionSpace}

Table~\ref{tab:obsSpace} shows the complete 20-dimensional observation vector for RL agents. Spatial observations (distances, angles) are relative to the agent's position and forward direction. Table~\ref{tab:actionSpace} shows the action space.

\begin{table}[H]
\centering
\caption{RL agent observation vector.}
\label{tab:obsSpace}
\begin{tabular}{cll}
\toprule
\textbf{\#} & \textbf{Observation} & \textbf{Range} \\
\midrule
1  & Goal distance       & $[0, 1]$ \\
2  & Goal angle          & $[-1, 1]$ \\
3  & In goal zone        & $\{0, 1\}$ \\
4  & Ally distance       & $[0, 1]$ or $-1$ \\
5  & Ally angle          & $[-1, 1]$ \\
6  & Pickup available    & $\{0, 1\}$ \\
7  & Pickup angle        & $[-1, 1]$ \\
8  & Pickup distance     & $[0, 1]$ or $-1$ \\
9  & Enemy visible       & $\{0, 1\}$ \\
10 & Enemy angle         & $[-1, 1]$ \\
11 & Enemy distance      & $[0, 1]$ or $-1$ \\
12 & Time remaining      & $[0, 1]$ \\
13 & Health              & $[0, 1]$ \\
14--20 & Ray cast distances (7 rays) & $[0, 1]$ \\
\bottomrule
\end{tabular}
\end{table}

\begin{table}[H]
\centering
\caption{RL agent action space.}
\label{tab:actionSpace}
\begin{tabular}{lll}
\toprule
\textbf{Action} & \textbf{Type} & \textbf{Range} \\
\midrule
Move forward / backward & Continuous & $[-1, 1]$ \\
Turn left / right       & Continuous & $[-1, 1]$ \\
Strafe left / right     & Continuous & $[-1, 1]$ \\
Jump                    & Discrete   & $\{0, 1\}$ \\
Fire weapon             & Discrete   & $\{0, 1\}$ \\
\bottomrule
\end{tabular}
\end{table}

\section{Reward Function Values}
\label{app:rewardValues}

Table~\ref{tab:rewardDesc} describes each reward component. Table~\ref{tab:rewardAll} lists the concrete weight values used for each skill policy. A dash indicates that the component is unused (zero weight) for that skill. The Default policy is used for the Flat RL baseline.

\begin{table}[H]
\centering
\caption{Reward components and their descriptions.}
\label{tab:rewardDesc}
\begin{tabularx}{\textwidth}{lX}
\toprule
\textbf{Component} & \textbf{Description} \\
\midrule
Distance shaping        & $(d_{t-1} - d_t) \times \text{scale}$: per-step reward proportional to the reduction in normalized distance to the target (goal zone or health pickup). \\
Zone bonus              & Constant reward per step while the agent is inside the goal zone. \\
Time penalty            & Small negative reward per step to discourage passive behavior. \\
Angle-to-enemy          & $(1 - \text{clamp}(\theta_{\text{enemy}})) \times \text{scale}$: reward for facing an enemy, scaled by alignment accuracy. \\
Damage dealt            & Reward per point of damage inflicted on an enemy. \\
Damage taken            & Penalty per point of damage received from an enemy. \\
Kill reward             & One-time reward when the agent eliminates an enemy. \\
Pickup reward           & One-time reward when collecting a health pickup. \\
Weapon fire             & Penalty per shot fired, suppressing indiscriminate shooting. \\
Jump                    & Penalty per jump, discouraging unnecessary jumping. \\
Terminal (success)      & One-time reward at episode end if the agent's team wins. \\
Terminal (failure)      & One-time penalty at episode end if the agent's team loses. \\
\bottomrule
\end{tabularx}
\end{table}

\begin{table}[H]
\centering
\caption{Reward function weight values for all skill policies.}
\label{tab:rewardAll}
\begin{tabular}{l rrrrr}
\toprule
\textbf{Component} & \textbf{Navigate} & \textbf{Combat} & \textbf{Secure} & \textbf{Retreat} & \textbf{Default} \\
\midrule
Distance shaping        & $0.1$    & ---      & ---      & $0.5$    & $0.05$  \\
Zone bonus              & $0.01$   & ---      & $0.05$   & ---      & $0.02$  \\
Time penalty            & $-0.01$  & ---      & $-0.01$  & $-0.01$  & $-0.002$ \\
Angle-to-enemy          & ---      & $0.005$  & $0.005$  & ---      & $0.005$ \\
Damage dealt            & ---      & $0.05$   & $0.05$   & ---      & $0.05$  \\
Damage taken            & $-0.05$  & $-0.025$ & $-0.025$ & $-0.02$  & $-0.05$ \\
Kill reward             & ---      & $3.0$    & $3.0$    & ---      & $3.0$   \\
Pickup reward           & ---      & ---      & ---      & $10.0$   & $1.5$   \\
Weapon fire             & $-0.05$  & $-0.005$ & $-0.005$ & $-0.02$  & $-0.01$ \\
Jump                    & $-0.02$  & $-0.01$  & $-0.01$  & $-0.01$  & $-0.02$ \\
Terminal (success)      & ---      & ---      & ---      & ---      & $10.0$  \\
Terminal (failure)      & ---      & ---      & ---      & ---      & $-1.0$  \\
\bottomrule
\end{tabular}
\end{table}

\section{PPO Training Hyperparameters}
\label{app:ppoHyperparams}

All policies use identical PPO hyperparameters during training, summarized in Table~\ref{tab:ppoHyperparams}. Training was performed using Unity ML-Agents~3.0.0 with 32 parallel environment instances. Each skill policy (Navigate, Combat, Secure, Retreat) was trained for 10 million steps, while the Default policy (used as the Flat RL baseline) was trained for 30 million steps.

\begin{table}[H]
\centering
\caption{PPO hyperparameters used for all policy training.}
\label{tab:ppoHyperparams}
\begin{tabular}{ll}
\toprule
\textbf{Parameter} & \textbf{Value} \\
\midrule
Batch size              & 64 \\
Buffer size             & 1\,200 \\
Learning rate           & $3 \times 10^{-4}$ (linear decay) \\
Entropy coefficient     & 0.001 \\
PPO clip ($\epsilon$)   & 0.2 \\
GAE $\lambda$           & 0.99 \\
Epochs per update       & 3 \\
Hidden units            & 128 \\
Hidden layers           & 2 \\
Observation normalization & Enabled \\
Discount factor ($\gamma$) & 0.99 \\
Time horizon            & 200 \\
\bottomrule
\end{tabular}
\end{table}

\section{LLM Prompt}
\label{app:llmPrompt}

The LLM controller receives a two-part prompt at each decision step: a static system prompt and a dynamic user prompt containing the current game state.

\subsection{System Prompt}
Listing~\ref{lst:systemPrompt} shows the complete system prompt, which encodes game rules, available skills with usage guidelines, team coordination heuristics, and output format.

\begin{lstlisting}[
  basicstyle=\ttfamily\small,
  breaklines=true,
  breakatwhitespace=true,
  frame=single,
  backgroundcolor=\color{gray!8},
  caption={System prompt provided to the LLM.},
  label={lst:systemPrompt},
  captionpos=b
]
You are a tactical coordinator for a 2v2 king of the hill game.
Goal: hold the goal zone for 8 seconds total to win.
Death penalty: lose 50% goal progress.

Available skills:
  1 = Navigate - Move toward goal zone
      USE: When far from goal zone, or when near goal but need to enter after clearing enemies
  2 = Combat - Fight enemies
      USE: Only when an enemy is visible. Especially near goal zone (clear threats BEFORE entering), or when providing perimeter defense while ally secures zone
  3 = Secure - Defend goal zone you already control
      USE: Only when inside goal zone
  4 = Retreat - Emergency escape to health pickup
      USE: Only when at critical health AND outside goal zone AND health pickup is available

Team coordination: Only one agent needs to be in zone to capture/hold. If ally is securing zone, other agent should provide perimeter defense with Combat.

IMPORTANT: Reply INSTANTLY with ONLY two numbers separated by comma (e.g., "1,3"). NO explanations, NO thinking, NO reasoning - just the numbers.
\end{lstlisting}

\subsection{User Prompt}
Listing~\ref{lst:userPromptExample} shows an example user prompt with populated state variables. State variables are discretized into categorical labels to enable semantic reasoning. Table~\ref{tab:llmObs} lists all observation fields and discretization thresholds.

\begin{lstlisting}[
  basicstyle=\ttfamily\small,
  breaklines=true,
  breakatwhitespace=true,
  frame=single,
  backgroundcolor=\color{gray!8},
  caption={Example user prompt with current game state.},
  label={lst:userPromptExample},
  captionpos=b
]
Current game observations:
Agent1 [Current: Secure]:
  Health: HIGH (100 %)
  Goal distance: NEAR (0.00)
  In goal zone: YES
  Enemy visible: YES (dist: NEAR)
  Health pickup: YES (dist: NEAR)

Agent2 [Current: Combat]:
  Health: MEDIUM (53 %)
  Goal distance: NEAR (0.01)
  In goal zone: NO
  Enemy visible: YES (dist: NEAR)
  Health pickup: YES (dist: NEAR)

Agent distance: NEAR
Time: PLENTY (80 %) | Goal progress: 60 %
\end{lstlisting}

\begin{table}[H]
\centering
\caption{Observation fields included in the LLM user prompt with discretization thresholds.}
\label{tab:llmObs}
\begin{tabular}{ll}
\toprule
\textbf{Field} & \textbf{Format / Thresholds} \\
\midrule
\multicolumn{2}{l}{\emph{Per-agent}} \\
Health              & HIGH ($>66\%$) / MEDIUM ($>33\%$) / LOW \\
Goal distance       & NEAR ($<0.15$) / MID ($<0.30$) / FAR \\
In goal zone        & YES / NO \\
Enemy visible       & YES / NO \\
Enemy distance      & NEAR ($<0.15$) / MID ($<0.30$) / FAR (if visible) \\
Health pickup       & YES / NO \\
Pickup distance     & NEAR ($<0.15$) / MID ($<0.30$) / FAR (if visible) \\
Current skill       & NAVIGATE / COMBAT / SECURE / RETREAT \\
\midrule
\multicolumn{2}{l}{\emph{Team-level}} \\
Agent distance      & NEAR ($<0.15$) / MID ($<0.30$) / FAR \\
Time remaining      & PLENTY ($>66\%$) / LIMITED ($>33\%$) / URGENT \\
Goal progress       & Percentage (0--100\%) \\
\bottomrule
\end{tabular}
\end{table}

\section{Agent-Level Metrics}
\label{app:agentLevelMetrics}
\begin{figure}[H]
    \centering
    \includegraphics[width=0.9\textwidth]{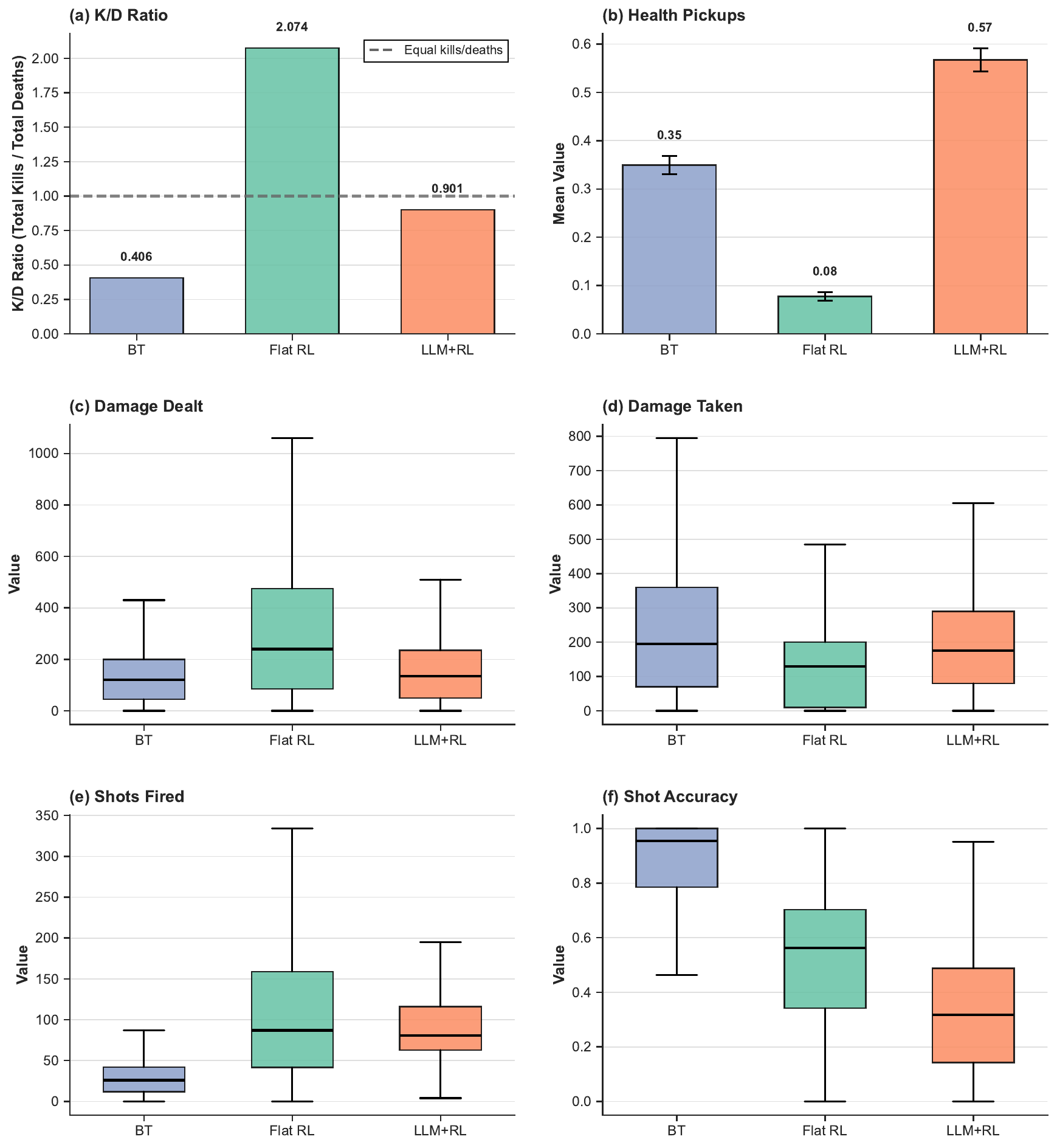}
    \caption{Agent-level metrics aggregated across all episodes ($n=4{,}000$ per type). (a) K/D Ratio with dashed line at 1.0. (b) Health pickups per episode with 95\% confidence intervals. (c--f) Combat behavior distributions.}
    \label{fig:agentMetricsAggregate}
\end{figure}

Figure~\ref{fig:agentMetricsAggregate} presents detailed agent-level metrics aggregated across all performance evaluation episodes ($n=4{,}000$ per agent type), including K/D ratios, health pickup collection rates, and combat behavior patterns (shots fired, accuracy, damage dealt/taken). 

\section{User Study: Participant Responses}
\label{app:userStudyResponses}

This appendix provides verbatim participant responses to open-ended questions from the user study. Participants ($n=15$) provided feedback after each block and completed a final comparative questionnaire. For reference, the block presentation order was: Block 1 = BT, Block 2 = Flat RL, Block 3 = LLM+RL. The order was counterbalanced across participants.

\subsection{Post-Block Observations}
\label{sec:postBlockObservations}

After each block, participants answered: \textit{"Did you notice any specific behaviors or patterns that influenced your impression?"}

\subsubsection*{Block 1 -- BT}

\begin{enumerate}[noitemsep]
\item "When they were about to die, they left the goal to pick up the item to recover their health."
\item "They all went to generate health at the same health percentage"
\item "Both on one spot all the time. Went for pick up (health) after a certain health amount."
\item "They always ran to get a health boost when the health bar went low"
\item "In general felt like they employed a similar strategy all rounds, but there was one time they sneaked up behind us when we were watching two different angles which felt very human"
\item "Lower precision compared to other two blocks"
\item "They seemed to focus on one thing at a time, i.e. only shooting. Or only moving. Sometimes they were standing as I shot them without reacting. This leads me to believe they were AI. On coordination, they did arrive two by two all the time, which felt somewhat coordinated."
\item "The AI rushed the point as if they immediately knew where it was from the start, whereas we the humans had to figure out where it was for a split second every time. They felt a bit too robotic in the sense that they were always faster than us to the goal. The combat itself wasn't too difficult either, since their main focus seemed to be on the objective. They did, however, act a bit coordinated since they could stop and shoot at you if they caught you alone outside the objective. They also had some small "tactics" at times, with one standing on the objective and the other shooting from the side."
\item "As soon as the round started they rushed to the capture point, they always knew where it was. They did not seem to utilize mechanics such as jump, and once they got low health, it seemed like they immediately went to get a health potion."
\item "They were going for health this time which was a bit more skilled move. They rushed towards the goal"
\item "The opponents seemed to get to the capture area faster which made them more challenging to beat. This also made them seem more AI than humans."
\item "They seemed to consistently retreat either when they were low health (To pickup health) or when they were alone perhaps. The biggest telltale sign of them being AI was how static and still they stood while on the objective, while other players would've jumped around and kept moving to through the opponents aim off. They also seemed to be consistent in their actions, repeating events such as retreating, etc., while humans might act differently in the same situation the second time."
\item "They seem to try and kill you as a priority before capturing the objective"
\item "They moved as a unit, and seemed to go for the health pickups when they were low. Not fast enough though."
\end{enumerate}

\subsubsection*{Block 2 -- Flat RL}

\begin{enumerate}[noitemsep]
\item "The opponents seemed to focus on attacking us rather than go for the objective if they saw us before seen the goal. This is similar to the behavior of human players in Battlefield. Also the opponents shot from afar which was different compared with the other blocks."
\item "They were better but had a pretty consistent tactic, more long range and focused. But no ``moments of genius''."
\item "AI, because random jumps, very good aim when standing next to them"
\item "Maybe it felt a bit more tactical by staying back a bit from the capture point (compared to just rushing in on it immediately), but it didn't feel like it gave them an advantage and they didn't adapt from this strategy"
\item "They used jumps to peek and shoot. They also sometimes used walls as cover. But still It feels that once they started shooting they didnt change their position a lot, which made it easier for me to hit them"
\item "Very good precision shooting me"
\item "Sometimes they were jumping which made it a bit challenging but in general but they didn't seem to move around a lot when I was targeting them. No signs of coordination"
\item "This time around they felt more like a challenge overall, since they were moving more, hitting more of their shots as well as not just B-lining the objective as they did previously. They felt more coordinated, where they often tried to move together and sometimes feeling a bit more tricky/sneaky by hiding behind walls. They still felt a bit robotic though because of their lack of variety in approach. Like, they used the same tactic of approaching the objective from a distance and shooting from outside of it every round to try and take us out before we get there."
\item "The AI seemed to play very passive, they did not challenge the capture point after we were on it."
\item "they were keeping their distance and coordinated a bit better. hard to get them because of distance but more challenging"
\item "The opponents didn't seem to move to much while shooting which made them easy targets."
\item "They seemed to be less eager to push the objective of both of them, opting to stay at a distance and jump randomly. It worked well to throw off my aim, but by never pushing the objective, they seemed like AI agents. Definitely caught me off guard once or twice though, so if they pushed the objective more I would've perhaps believed they were human players."
\item "They really prioritized sniping me before ever pushing to the objective"
\item "They worked as a unit with one in the front and one in the back. they also seemed to focus their fire on one player, which meant we died more often"
\end{enumerate}

\subsubsection*{Block 3 -- LLM+RL}

\begin{enumerate}[noitemsep]
\item "It seemed like one of the opponents was always on the outside of the protecting area, while the other was inside trying to hold the objective. Also, one of the opponents tried to protect itself by jumping and sometimes it ended up on top of trees. I am not sure if this was intentional or not."
\item "Too bad to be human. But they lowkey had some tactics. They had one take spot and one snipe long range, but used that tactic pretty consistently."
\item "Placing a bit random"
\item "I think it felt a bit more dynamic, maybe, somewhat better teamwork and while the strategy used felt pretty similar between turns it felt more well-executed and on occasion they'd adapt to what was going on in the round"
\item "They used the environment very well. Like jumping over walls to get to the objective faster. Also one of them jumped on a pillar while the other one was hiding behind it. They also stood on the health spot spawn to refill as soon as it was available."
\item "more jumps, distracted me quite often"
\item "In 5 of 6 games they coordinated well arriving two at a time. They were active, jumping around or moving to avoid getting hit. Also picking up health. Could have been human for sure but also good AI."
\item "This time around the AI felt more like a challenge as they also managed to win some rounds against us. I believe they still felt a bit robotic by being very quick to get to the point every round, but overall they felt more human by being a bit unpredictable. Both by jumping around to dodge bullets and not always just standing around shooting at you. But also because they would approach you and the objective a bit differently from round to round."
\item "They were aggressive and challenged the capture point they also utilized mechanics such as jump."
\item "Moved straight to the goal and jumping like crazy. rotating a bit odd"
\item "The opponents were going for health pick ups when they had low HP which made them seem more human."
\item "The erratic jumping made them seem like AI, since it wasn't really making sense. Just jumping on the spot sometimes. Also one of the AIs stayed for a long time on their spawn, which seemed like an AI malfunction. They did feel unpredictable sometimes, acting illogically, which made my judgment lean a little into the human since they might be terrible human gamers."
\item "The enemy just b-line so I shoot them and do the same"
\item "They were attacking one at a time, which seems not human-like when you're on the offensive"
\end{enumerate}

\subsection{Final Observation}
\label{sec:forcedChoiceExplanations}

After experiencing all three agent types, participants selected which team felt most like competent human players and explained their choice.

\subsubsection*{LLM+RL Selected}

\begin{enumerate}[noitemsep]
\item "Because they seemed to match how I played the game. Utilize all the mechanics, did not immediately know where the site was it felt like, they were not too aggressive but also aggressive enough."
\item "Peeking, using the environment."
\item "Multiple actions at the same time. Jumping and shooting while they were behind walls which feels quite human."
\item "There were a lot of kinds of human behaviors such as the random jumping (which I didn't notice with the other teams, but not sure if that's the case), how aggressive they were (in my experience usually more of a human element than an AI element in video games). Also, in general, they played better, and it felt like there were more close calls where they could've won."
\item "Any time I got the feeling that they adapted their behavior to mine it felt more human-like, when they always execute the same behavior it feels like a bot in contrast"
\item "More movement and jumping and using the health pick ups when low on health."
\item "I saw no direct pattern of how the players behaved, unlike Team 1"
\item "Going or not for health bag when low HP, jump or not to disturb us"
\end{enumerate}

\subsubsection*{Flat RL Selected}

\begin{enumerate}[noitemsep]
\item "They were better. And the tactic seemed smartest, leveraging range to prevent us from getting the flag."
\item "The way the second block maneuvered the level and moved during combat felt the most realistic and challenging. Jumping around and strafing. Although as mentioned, them almost never being on objective didn't seem very human. A human would've rushed in, even during the risk of dying once or twice."
\item "Cooperation, focusing fire. Seemed they had a strategy to eliminate players rather than just heading dead-on to the zone"
\item "They kept the distance and did not rush as fast as the other ones"
\item "Focusing on damaging the opponent before going to secure the objective."
\end{enumerate}

\subsubsection*{BT Selected}

\begin{enumerate}[noitemsep]
\item "They were moving more like humans, like me and my teammate moving around the circle, searching for health, not jumping all over the place"
\end{enumerate}

\subsection{General Feedback}
\label{sec:generalFeedback}

Optional final question: \textit{"Any other thoughts about the opponents or gameplay?"}

\begin{enumerate}[noitemsep]
\item "The controls felt too fast for me."
\item "The shots could be neon green or something to make the shots more visible. The players look cute."
\item "Fun game, but need scope to aim better"
\item "General gameplay: getting stuck in the corner of objects and trees. Jumping didn't have a speed penalty, so it's worth mentioning this to the player at the start. The objective sign was not always visible."
\item "I would want some more game feedback on if you're getting hit or hitting an opponent with the bullets, also the bullets are barely visible which can be improved. With those changes, it will be easier to observe behavior in a pvp shooter"
\item "Interesting simulation! Maybe reloading the weapon could further enhance shooting/staying in cover mechanics. Also I wanted to jump with the A button"
\item "Agents: I did not see an example of opponents attacking you from both sides, but I saw them jumping and moving out of the line of sight in block 3."
\item "Don't feel much difference about collaboration, they rarely attack together on one target"
\item "Would be nice if they coordinated between number 1 and 2. So they could both retreat and shoot from far but also get health and one could go for the zone. So, combining them all would feel maybe more human-like."
\end{enumerate}

\section{Skill Analysis}
\label{app:llmAnalysis}
\begin{figure}[ht]
    \centering
    \includegraphics[width=\textwidth]{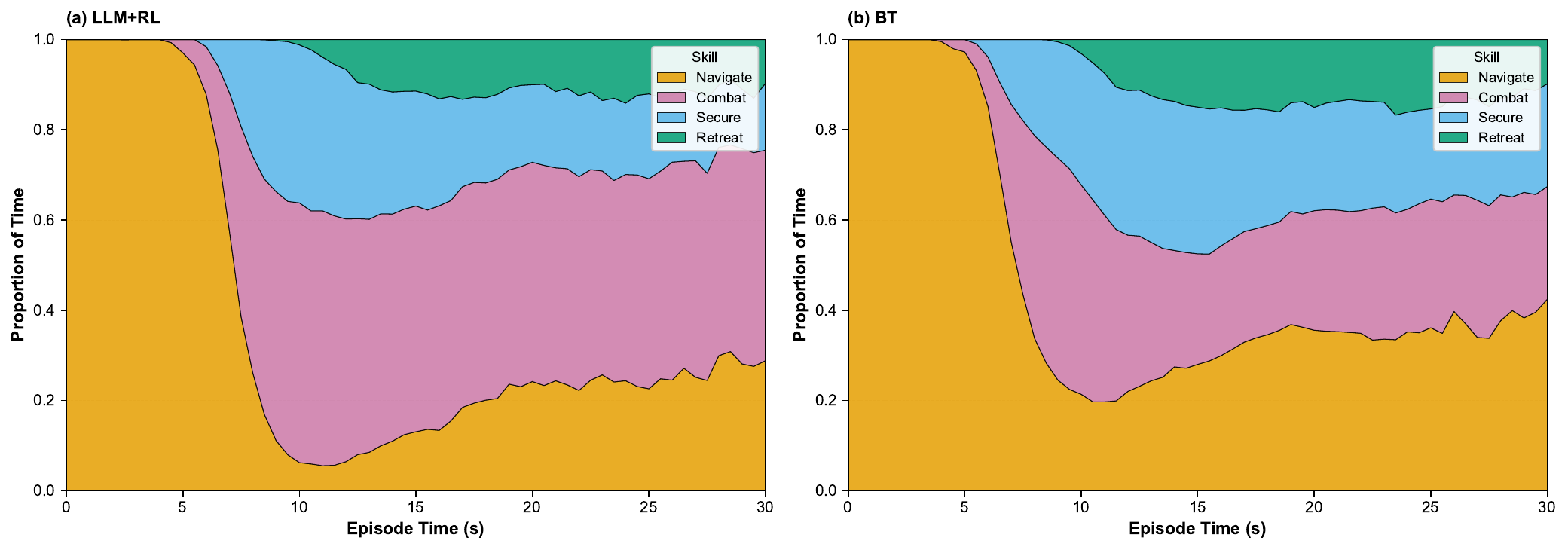}
    \caption{Skill usage over episode time. Both agents show strategic phase adaptation: Navigate dominates early, Combat and Secure increase mid-episode, and Retreat emerges late. LLM+RL exhibits a more decisive shift from Navigate to Combat/Secure around 10 seconds.}
    \label{fig:skillOverTime}
\end{figure}

\begin{figure}[ht]
    \centering
    \includegraphics[width=\textwidth]{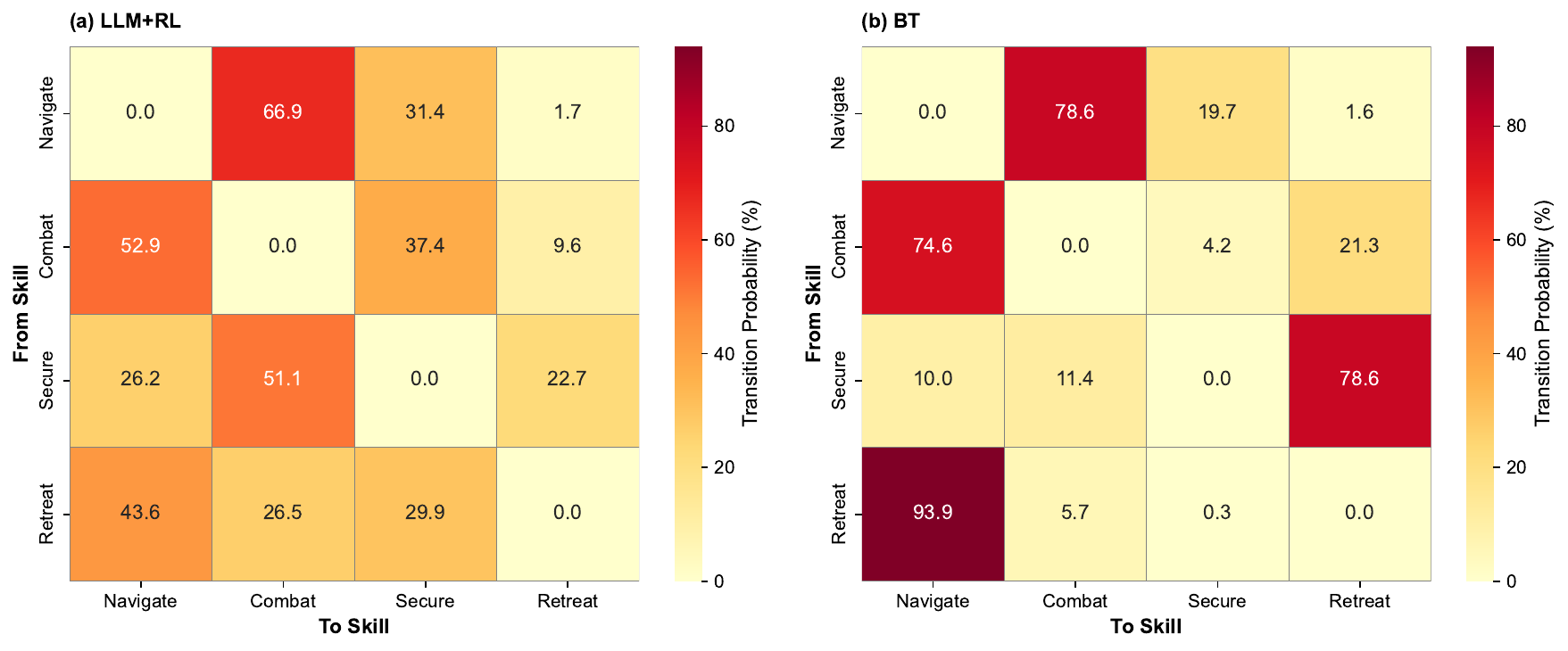}
    \caption{Skill transition matrices showing from-skill (rows) to to-skill (columns) probabilities. Rows represent row-wise conditional probabilities and sum to 100\%. LLM+RL exhibits higher transition entropy ($H=3.19$ bits vs $H=2.47$ bits for BT), indicating more diverse transitions rather than stereotyped loops.}
    \label{fig:skillTransitions}
\end{figure}

\begin{figure}[ht]
    \centering
    \includegraphics[width=\textwidth]{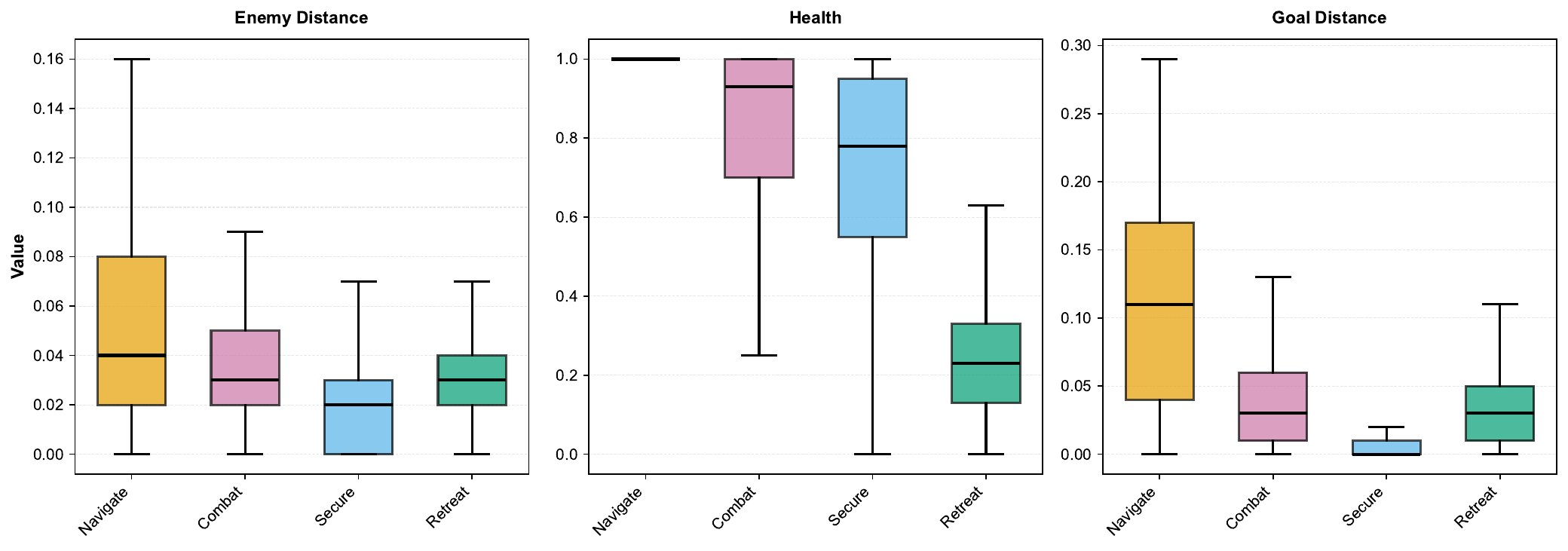}
    \caption{Skill selection patterns by game-state context for LLM+RL. Strong correlations emerge: Retreat dominates at low health, and Secure when in goal zone, demonstrating context-dependent skill selection.}
    \label{fig:decisionContext}
\end{figure}

Figure~\ref{fig:skillOverTime} shows temporal skill usage patterns, Figure~\ref{fig:skillTransitions} presents skill transition matrices with entropy values, and Figure~\ref{fig:decisionContext} illustrates how skill selections correlate with game-state context for LLM+RL.

\end{document}